\pdfoutput=1

\documentclass[11pt]{article}

\usepackage[review]{acl}

\usepackage{times}
\usepackage{latexsym}

\usepackage[T1]{fontenc}

\usepackage[utf8]{inputenc}

\usepackage{microtype}

\usepackage{graphicx}
\usepackage{multirow}
\usepackage{amsmath}

\usepackage{epstopdf}
\title{MPN: Leveraging Multilingual Patch Neuron for Cross-lingual Model Editing}

\author{Nianwen Si$^{1,2}$, Hao Zhang$^1$, Weiqiang Zhang$^2$ \\
$^1$Information Engineering University, Zhengzhou, China \\
$^2$Department of Electronic Engineering, Tsinghua University, Beijing, China}
\date{}

\begin{document}
\maketitle
\begin{abstract}
Large language models are known for encoding a vast amount of factual knowledge, but they often becomes outdated due to the ever-changing nature of external information. A promising solution to this challenge is the utilization of model editing methods to update the knowledge in an efficient manner. However, the majority of existing model editing techniques are limited to monolingual frameworks, thus failing to address the crucial issue of cross-lingual knowledge synchronization for multilingual models. To tackle this problem, we propose a simple yet effective method that trains multilingual patch neuron to store cross-lingual knowledge. It can be easily adapted to existing approaches to enhance their cross-lingual editing capabilities. To evaluate our method, we conduct experiments using both the XNLI dataset and a self-constructed XFEVER dataset. Experimental results demonstrate that our proposed method achieves improved performance in cross-lingual editing tasks without requiring excessive modifications to the original methodology, thereby showcasing its user-friendly characteristics. Codes will be released soon.
\end{abstract}

\section{Introduction}

Large language models (LLMs) have made remarkable progress and become a forefront technology in natural language processing recently. With their extensive factual knowledge learnt during pretraining and fine-tuning, LLMs are often compared to knowledge bases. However, the presence of erroneous knowledge in training data and the ever-changing nature of real-world information can result in issues such as knowledge truncation and errors within LLMs. Therefore, it is crucial to regularly modify and update the model's knowledge to address these challenges.

One natural approach to updating a model’s knowledge is to retrain it on new corpora. Unfortunately, this method has significant computational costs, as LLMs require extensive resources, including compute power and time. An advanced approach is parameter-efficient fine-tuning, but it has the risk of overfitting and catastrophic forgetting. More recently, researchers have introduced model editing methods that allow for calibrating specific knowledge without affecting other aspects of the model, showing promise in addressing these challenges. These methods can be categorized into three groups: 1) Fine-tuning based methods, which fine-tune some model parameters to encode specific knowledge while constraining changes to unrelated knowledge; 2) Hypernetwork based methods, which use a hypernetwork to predict the parameter changes needed for specific inputs; and 3) Locate-then-edit methods, which identify neurons related to specific knowledge and fine-tune them to modify their semantic expression.

\begin{figure}
    \centering
    \includegraphics[width=1\linewidth]{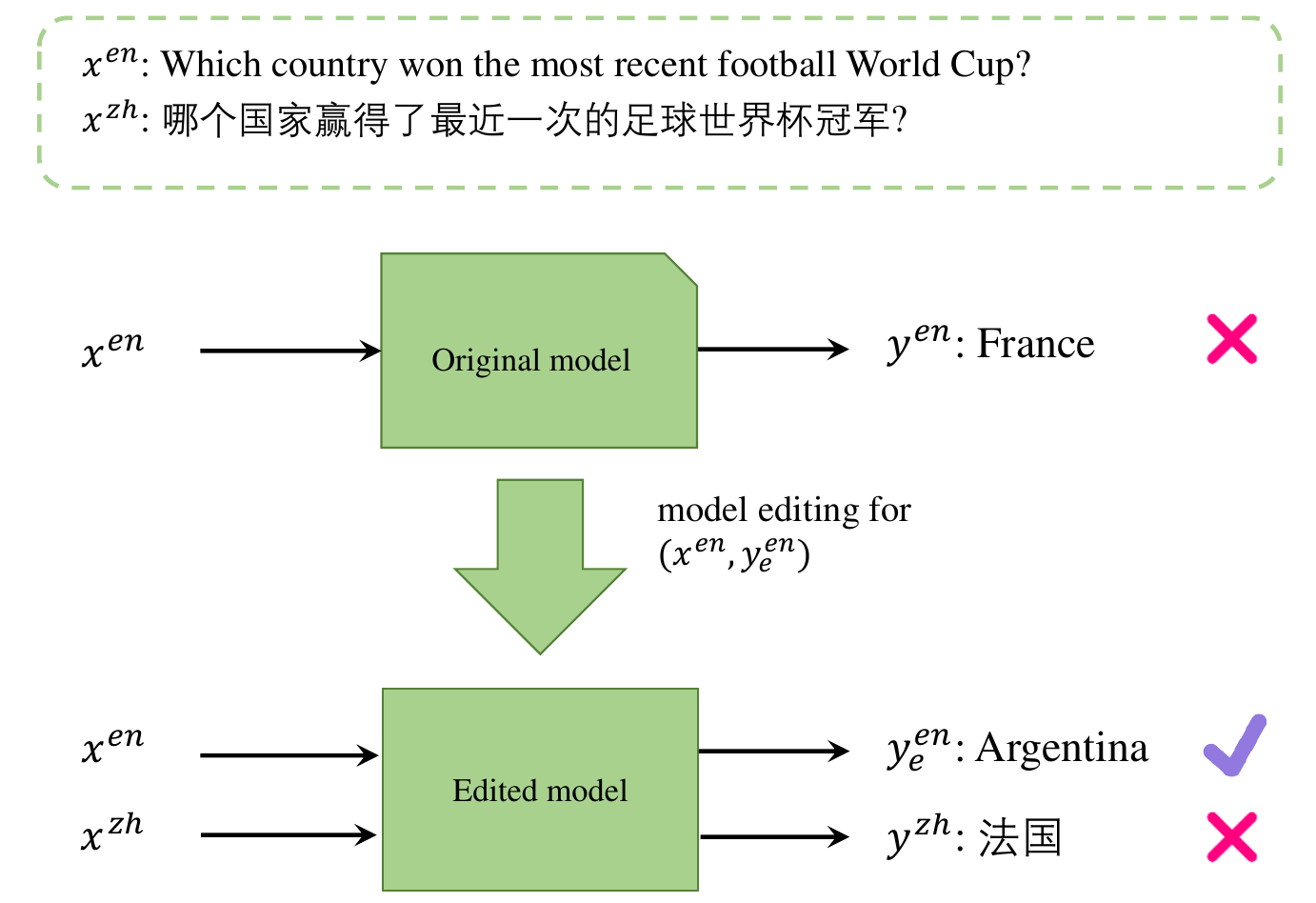}
    \caption{Monolingual model editing fails in cross-lingual scenario}
\end{figure}

However, these methods primarily focus on monolingual editing and overlook the cross-lingual transferability of the editing results. For example, when editing a model with English examples, the edited model may provide updated responses in English but still offer outdated answers in other languages, as shown in Figure 1. Since most large models encode knowledge in multiple languages, cross-lingual editing is a highly relevant problem that necessitates simultaneous updating knowledge across multiple languages.

Currently, only a few works have explored the cross-lingual effects of model editing. Bi-zsRE \cite{wang2023cross} assessed the cross-lingual capabilities of existing monolingual editing methods using English and Chinese datasets. The results showed that the cross-lingual generalizability of existing methods was unsatisfactory. LiME \cite{xu2022language} leveraged the language anisotropic characteristics of multilingual models to adaptively learn a language-specific mask for methods like KnowledgeEditor \cite{de2021editing} and MEND \cite{mitchell2021fast} using hypernetworks. This approach improved cross-lingual editing effectiveness, but it requires modifications to the editing method and is limited to hypernetwork-based methods. AMIG \cite{chen2023journey} focus solely on English-Chinese editing and ignore the generalization to other languages. It follows a locate-then-edit way by identifying knowledge neurons using multilingual integrated gradients \cite{sundararajan2017axiomatic} and then editing them. However, other study reveals that updating knowledge by identifying and editing neurons has limited cross-lingual effectiveness due to the inconsistency of multilingual knowledge in models \cite{qi2023cross} .

The challenge of cross-lingual model editing lies in the language gap that exists between different languages. This means that when editing a model for one language, the same fact described in another language may not change as well. Previous researches on cross-lingual transfer in multilingual models have struggled with this issue. Explicit multilingual fine-tuning methods, such as tuning the language adapters for various languages with multilingual datasets, have been used to bridge the language gap \cite{pfeiffer2020mad,parovic2022bad,parovic2023cross}. Inspired by this, this paper proposes the multilingual patch neuron (MPN), a straightforward yet effective method for editing multilingual model. It only needs to modify the data sampler during the training process of the existing editing method. By using English examples alongside any parallel corpus as input, the MPN can greatly improve the cross-lingual ability of the existing monolingual editing method without extra modifications of the editing method itself. Up to now, our work is among the few studies to explore the cross-lingual generalization of model editing. 

The main contributions of this paper are as follows: 

\begin{itemize}
\item We find that existing fine-tuning based editing methods possess a certain degree of cross-lingual editing effect. This effect is not derived from the editing method itself, but rather from the inherent cross-lingual transfer ability of the multilingual model.
\item We introduce a straightforward yet practical cross-lingual editing method named MPN based on Transformer-patcher \cite{huang2023transformer}, which has good versatility and can be extended to other fine-tuning based editing methods. Without increasing any complexity, it endows the monolingual editing methods with the ability to perform cross-lingual editing while maintaining the original effectiveness.
\item We demonstrate the cross-lingual ability of MPN on two multilingual datasets, XFEVER and XNLI. XFEVER is a multilingual fact-checking dataset we created based on FEVER dataset, containing six languages of three different language families. The results show that our method can enhance the efficacy of cross-lingual editing, with an average improvement of about 9\% in the five languages of XFEVER dataset and about 12\% in the 14 languages of XNLI dataset.
\end{itemize}

\section{Related work}
\subsection{Model Editing}
Model editing is an efficient knowledge calibrating method for LLMs that appears in recent years. Current model editing methods can be divided into three main categories:

\textbf{Methods based on fine-tuning}. The simplest way is to fine-tune the whole model directly on the new knowledge, but this will inevitably affect the existing knowledge of the model and cause catastrophic forgetting. Research on the internal representation of transformer shows that the feed-forward neural network (FFN) layer in transformer block is the main location to encode factual knowledge. It can be explicitly expressed in the form of key-value pairs similar with self-attention mechanism, where key represents the first linear layer and value represents the second linear layer of FFN \cite{geva2020transformer,dai2021knowledge}. Building upon this, a variety of model editing methods based on fine-tuning FNN layer are proposed, such as knowledge neuron \cite{dai2021knowledge}, CaliNet \cite{dong2022calibrating}, GRACE \cite{hartvigsen2022aging}, and Transformer-patcher \cite{huang2023transformer}. By fine-tuning a sum number of additional parameters added to the model, it is able to learn new knowledge and minimize the impact on existing knowledge. In addition, several locate-then-edit methods have been proposed, including ROME \cite{meng2022locating}, MEMIT \cite{meng2022mass}, and PMET \cite{li2023pmet}. These methods involve identifying key neurons associated with particular pieces of knowledge and then adjusting these neurons as necessary to modify or update the factual knowledge stored within the model.

\textbf{Methods based on hypernetwork}. The so-called hypernetwork refers to an additional network used to predict the updates of the parameters in target network. The main idea of hypernetwork based method involves training an editor (hypernetwork) using a dataset of new knowledge, allowing it to learn how to edit knowledge and maintain generalization and locality of new knowledge. When providing new facts, the trained editor can automatically predict the necessary parameter updates for editing these facts. Examples of methods based on hypernetwork include KnowledgeEditor \cite{de2021editing}, MEND \cite{mitchell2021fast}, and MALMEN\cite{Tan2023Massive}. KnowledgeEditor employs an LSTM as the editor, while MEND utilizes tensor decomposition to reduce the parameter size of the editor network, thereby simplifying the training process. MALMEN formulates the parameter shift aggregation as the least square problem, subsequently updating the model parameters using the normal equation.

\textbf{Methods based on in-context learning}. Methods such as IKE \cite{zheng2023can} and MemPrompt \cite{madaan2022memory} treat the model as a black box and leveraged its few-shot prompt learning ability to rectify knowledge. In this approach, the model is prompted using the input question, and new knowledge is provided alongside as a demonstration. These methods rely on the large model’s excellent reasoning ability and perform well in the inferential generalization of editing results. For example, models like LLaMA-2 \cite{touvron2023llama} process strong chain-of-thought reasoning ability, enabling them to decompose multi-hop question into sub-problems with answers during the inference process. However, if the format of sub-question decomposition is inconsistent with expectations, the editing effect will be limited.  

In addition, several other methods are proposed like SERAC \cite{mitchell2022memory} and MELO \cite{Yu2023MELO}. Despite the numerous proposed editing methods, several challenging issues in this field are not well solved, such as cross-lingual generalization, reasoning on editing results, and sustainable editing. This paper mainly focuses on the cross-lingual editing problem.

\subsection{Cross-lingual Transfer}
Cross-lingual transfer refers to the ability of a model trained on one language to improve its performance on corresponding tasks in another language. Multilingual models possess cross-lingual transfer ability due to their implicit unified knowledge representation across different languages \cite{wu2019emerging}. However, this often results in a “transfer gap” where the task performance in the target language is generally worse than that in the source language. In traditional cross-lingual transfer methods, the common approach is to fine-tune the model directly on the source language for downstream tasks and then test it on the task set of the target language. In this case, low-resource languages and languages that are far from the source language tend to have poorer transfer performance on downstream tasks, resulting in a more pronounced transfer gap. Currently, The adapter is the mainstream method for cross-lingual transfer, with language-specific adapter and task adapter being two commonly used adapters that enable the model to adapt to both language and task \cite{pfeiffer2020mad,parovic2022bad,parovic2023cross}.

To make model editing methods also capable of cross-lingual capabilities, LiME \cite{xu2022language} proposes a language anisotropy model editing method and tests its effectiveness on XNLI \cite{conneau2018xnli} and mLAMA datasets \cite{kassner2021multilingual}. However, this method is only applicable to hypernetwork-based editing methods and requires modifications to the details of the editing method. AMIG \cite{chen2023journey} introduces a language-independent knowledge neuron method, considering only the generalization between English and Chinese. However, multilingual models like mBERT \cite{devlin2018bert} typically contain over 100 languages, thus this method cannot cover most languages in large models. Bi-zsRE \cite{wang2023cross} translates the zsRE dataset \cite{de2021editing,levy2017zero} into Chinese with the API interfaces of gpt-3.5-turbo and gpt-4 to create a bilingual version of Chinese-English. Then, it tests the mutual transfer effects of existing seven editing methods between English editing and Chinese editing. This work also has limited coverage of language types and does not investigate how to improve cross-lingual editing. 

\section{Cross-lingual Model Editing }
\subsection{Task Definition}
Model editing aims to updating the knowledge stored within a model, allowing for the acquisition of new knowledge without altering the original knowledge. During this process, three fundamental requirements need to be satisfied: reliability, generality, and locality \cite{yao2023editing}. Formally, given the original model $f(.;\theta)$ to be edit, where $\theta$ denotes the model parameters. The goal is to find a new model $f(.;\theta^{\prime})$ that satisfies reliability based on $f(.;\theta)$, while also ensuring generality and locality.

\textbf{Reliability}. Reliability refers to the successful editing of an example. For an example to be edited $(x_{edit}^{en},y_{edit}^{en})$, where $x_{edit}^{en}$ represents the input in English description and $y_{edit}^{en}$ represents the corresponding output, reliability requires the edited model $f(.;\theta^{\prime})$ to meet the following criteria:  
\begin{equation}\label{eq1}
f(x_{edit}^{en};\theta^{\prime})=y_{edit}^{en}
\end{equation}

In classification tasks, $y_{edit}^{en}$ is a classification label. In question-answering tasks, $y_{edit}^{en}$ denotes an output answer. Reliability is typically measured using the task loss during the model's training. For classification models, cross-entropy loss or KL loss is commonly used, e.g., $CrossEntropy(f(x_{edit}^{en};\theta),y_{edit}^{en})$.

\textbf{Generality}. In monolingual editing, generality requires that the edited model can accurately predict semantically rephrased examples. For the rephrased example $(x_{rephrase}^{en},y_{edit}^{en})\in R(x_{edit}^{en},y_{edit}^{en})$, the following are needed:
\begin{equation}\label{eq2}
f(x_{rephrase}^{en};\theta^{'})=y_{edit}^{en},\quad\forall x_{rephrase}^{en}\in R
\end{equation}

While for cross-lingual editing, generality should be extended to parallel expression in different languages in order to achieve cross-lingual generalizability. For example, the model edited on English data needs to consider not only generalize to English rephrases but also to cross-lingual parallel data. Therefore, for parallel example $(x_{parallel}^l,y_{edit}^l)\in P(x_{edit}^{en},y_{edit}^{en})$, where \textit{l} denotes the any other language expect English, the following are needed:
\begin{equation}\label{eq3}
f(x_{parallel}^l;\theta^{\prime})=y_{edit}^{en},\quad\forall x_{rephrase}^l\in P
\end{equation}

\textbf{Locality}. Locality denotes that the newly added knowledge does not affect the original knowledge stored in the model. To achieve locality, for unrelated example $(x^{\prime},y^{\prime})\in L(x_{edit}^{en},y_{edit}^{en})$, where $(x^{\prime},y^{\prime})$ denotes the semantically unrelated example to $(x_{edit}^{en},y_{edit}^{en})$, the following are needed:
\begin{equation}\label{eq4}
f(x^{\prime};\theta^{\prime})=y^{\prime},\forall x^{\prime}\in L
\end{equation}

LiME \cite{xu2022language} defines locality as irrelevant knowledge across languages. However, the storage of knowledge of different languages in multilingual models is not balanced, and a certain piece of knowledge encoded in a high-resource language not necessarily exist in a low-resource language. Since English examples are primarily used for editing in this work, we only consider irrelevant knowledge from English corpora and not from other language.

\subsection{Monolingual Baseline}
In the realm of cross-lingual transfer learning for multilingual models, fine-tuning stands out as the most effective methodology, which entails augmenting the training process with additional language-specific adapters to facilitate multilingual training \cite{pfeiffer2020mad,parovic2022bad,parovic2023cross}. When it comes to model editing, fine-tuning is also the most direct approach to acquiring new knowledge by adjusting certain parameters within the model. Transformer-patcher is one of these methods that add additional neuron for editing new knowledge sequentially into the model \cite{huang2023transformer}.

Transformer-patcher fine-tunes the final FFN layer of a model, treating this layer as a key-value pair and add new pairs to store new knowledge. It emphasizes sequential and sustainable editing, learning only one piece of knowledge at a time. For each new piece of knowledge, a new key-value pair (i.e. patch) is created and physically separated from the neurons of the original FFN neuron to minimize interference with the existing knowledge. The structure of the FFN layer with the added patch neuron is depicted in Figure 2.

\begin{figure}
    \centering
    \includegraphics[width=1\linewidth]{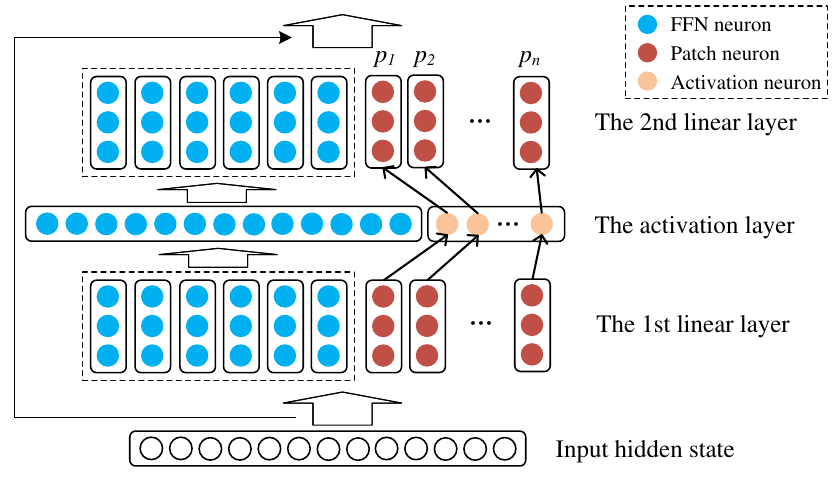}
    \caption{The patch neuron in FFN layer}
\end{figure}

From the perspective of key and value pair of knowledge neuron, the first and second linear layer of the FFN layer can be denoted as $\textbf{\textit{K}}$ and $\textbf{\textit{V}}$, respectively \cite{geva2020transformer,dai2021knowledge}. Given the input vector $\textbf{\textit{q}}$ to the FFN layer, the output $\mathrm{FFN}(\textbf{\textit{q}})$ can be calculated as follows: 
\begin{equation}\label{eq5}
\textbf{\textit{a}}=\sigma(\textbf{\textit{q}}\cdot \textbf{\textit{K}}+\textbf{\textit{b}}_\textbf{\textit{k}})
\end{equation}
\begin{equation}\label{eq6}
\mathrm{FFN}(\textbf{\textit{q}})=\textbf{\textit{a}}\cdot \textbf{\textit{V}}+\textbf{\textit{b}}_\textbf{\textit{v}}
\end{equation}
while $\sigma$ denotes the activation function, e.g. GLUE. $a$ is the activation of the first linear layer. For each example to be edited, a key-value pair will be added alongside the last FFN layer. Specifically, $\textbf{\textit{k}}_\textbf{\textit{p}}$ and $b_p$ is added with the first linear layer, and $\textbf{\textit{v}}_\textbf{\textit{p}}$ is added with the second linear layer. Then, the new forward process of the updated FFN layer can be formalized as follows:
\begin{equation}\label{eq7}
[\textbf{\textit{a}},a_p]=\sigma(\textbf{\textit{q}}\cdot[\textbf{\textit{K}},\textbf{\textit{k}}_\textbf{\textit{p}}]+[\textbf{\textit{b}}_\textbf{\textit{k}},b_p])
\end{equation}
\begin{equation}\label{eq8}
\mathrm{FFN}^{\prime}(\boldsymbol{\textbf{\textit{q}}})=[\textbf{\textit{a}},a_p]\cdot\begin{bmatrix}\boldsymbol{\textbf{\textit{V}}}\\\boldsymbol{\textbf{\textit{v}}}_\textbf{\textit{p}}\end{bmatrix}+\textbf{\textit{b}}_\textbf{\textit{v}}=\mathrm{FFN}(\boldsymbol{\textbf{\textit{q}}})+a_p\cdot\boldsymbol{\textbf{\textit{v}}}_\textbf{\textit{p}}
\end{equation}

Finally, the output $\mathrm{FFN}^{\prime}(\textbf{\textit{q}})$ is obtained which contains the calibration because of the added $a_p\cdot\boldsymbol{\textbf{\textit{v}}}_\textbf{\textit{p}}$. To ensure efficacy, the patch neuron should be maximally activated by the example to be edited. While for locality, memory loss on unrelated examples sampled from the training and test set is utilized. 

\subsection{Multilingual Patch Neuron}  
To improve cross-lingual editing effectiveness, we propose to use English examples as the primary input combined with any other parallel corpora to train multilingual patch neuron (i.e. MPN), enabling the patch to obtain the cross-lingual transfer ability. The reasons behind this is twofold: Firstly, while various languages in a multilingual model share the same semantic representation space, their abilities differ significantly. Improving the low-resource language ability of the model through training with high-resource languages is the most common method for achieving cross-lingual transfer. Therefore, it is essential to employ English examples during the training of patch neurons as English serves as a central language that maximizes the cross-lingual capability of the patch, and better drives the simultaneous update of knowledge in low-resource languages. Secondly, pairing English training with parallel examples of any other language can further enhance linguistic diversity in the editing process, resulting in multilingual patch neurons that allow to generalize better to other languages.

Specifically, for $(x_{edit}^{en},y_{edit}^{en})$ to be edited, a parallel example $(x_{edit}^{l},y_{edit}^{l})$ in another language $l\in language\_list$ is randomly sampled, and both examples are used for training the patch neuron, which can enhance the cross-lingual generalization ability of the patch. The loss function is formalized as following:
\begin{equation}\label{eq9}
\begin{split}
loss=CrossEntropy(f(x_{edit}^{en},x_{edit}^{l};\theta^{\prime}), \\
y_{edit}^{en}, y_{edit}^{l})
\end{split}
\end{equation}

MPN only requires modifying the input part of the Transformer-patcher without altering other components, which improves the cross-lingual generalization of the patch in a simple way. MPN differs from LiME in that LiME requires complete correspondence between parallel corpora of different languages for training, whereas we use weaker parallel corpora with less stringent requirements. Additionally, since only the input needs to be modified, MPN can be easily extended to other fine-tuning based editing methods to improve their cross-lingual editing effectiveness.

\section{Experiments}
\subsection{Settings}
\subsubsection{Models and Datasets}
We use bert-base-multilingual-uncased from Huggingface as the target multilingual model, which comprises 12 transformer blocks and supports 102 languages. Following the default settings of Transformer-patcher \cite{huang2023transformer}, we edit the final FFN layer of the model.

We conduct experiments on knowledge-intensive datasets, including FEVER and XNLI \cite{huang2023transformer,de2021editing}. For FEVER dataset, we follow previous work by splitting its training set into a ratio of 0.8:0.1:0.1 for training set $D_{train}$, validation set $D_{val}$, and editing set $D_{edit}$. This results in 83,972/10,496/10,496 examples, respectively. We use examples in $D_{edit}$ to edit the target model. The original development set is retained as the test set $D_{test}$ for testing locality, containing 10,444 examples. To perform cross-lingual editing, we use machine translation models to expand $D_{edit}$ into multiple language versions $D_{parallel}$, including German (de), French (fr), Spanish (es), Chinese (zh), and Arabic (ar). The final dataset is denoted as XFEVER. For details of XFEVER dataset and training the original model, please refer to Appendix A.

For XNLI dataset, it is a multilingual natural language inference dataset containing parallel corpora in 15 languages. Each sample comprises a premise and a hypothesis, and the task is to determine their entailment relationship: \{entailment, neutral, contradiction\}. It includes both high-resource languages like English (en) and low-resource languages like Swahili (sw), allowing for testing the transferability of editing methods for low-resource languages. We train the model using multiNLI dataset \cite{williams2017broad} containing 392,702 training examples. The development and test sets of XNLI dataset contain 2,490 and 5,010 examples for each language. The development set is used to determine the optimal trained model, and the test set is used to edit the model.

\subsubsection{Evaluation Metrics}
For reliability, it refers to the success rate on editing set $D_{edit}$ , which is defined as: 
\begin{equation}\label{eq10}
\mathrm{Reliability}=E_{(x,y)\sim D_{edit}}[I(f(x;\theta^{\prime})=y)],
\end{equation}
where $I$ is the indicator function.

For generality, we use the English rephrase set $D_{rephrase}$ from XFEVER dataset  to examine monolingual generalizaiton (MLG). While for XNLI dataset, we do not calculate the MLG since it does not contain rephrase data. MLG is formally defined as:
\begin{equation}\label{eq11}
\mathrm{MLG}=E_{(x,y)\sim D_{rephrase}}[I(f_t(x;\theta^{\prime})=y)]
\end{equation}

To examine cross-lingual generalization (CLG), we use the parallel set $D_{parallel}$ and CLG is defined as:
\begin{equation}\label{eq12}
\mathrm{CLG}=E_{(x,y)\sim D_{parallel}}[I(f_t(x;\theta^{\prime})=y)]
\end{equation}

For locality, following Transformer-patcher, a proportion of $D_{train}$ and all test set $D_{test}$ are used. For example, the locality on $D_{test}$ is defined as:
\begin{equation}\label{eq13}
\mathrm{Locality}_{\mathrm{test}}=E_{(x,y)\sim D_{test}}[I(f_t(x;\theta^{\prime})=y)]
\end{equation}

\begin{table*}
\caption{Results on XFEVER dataset}
\resizebox{\linewidth}{!}{
\begin{tabular}{cccccccccccc}
\hline
\multirow{2}{*}{Models/Methods} &
  \multirow{2}{*}{\begin{tabular}[c]{@{}c@{}}Patch\\ Num.\end{tabular}} &
  \multirow{2}{*}{\begin{tabular}[c]{@{}c@{}}Reliability\\ (en)\end{tabular}} &
  \multirow{2}{*}{MLG} &
  \multicolumn{6}{c}{CLG} &
  \multicolumn{2}{c}{Locality} \\ \cline{5-12} 
               &      &       &                & de    & fr    & es    & zh    & ar    & Avg.  & Train          & Test           \\ \hline
Original model & 0    & 88.07 & 84.86          & 85.32 & 85.92 & 85.96 & 82.73 & 81.69 & 84.32 & 93.05          & 77.08          \\ \hline
Fine-tuning    & 0    & 83.72 & 78.65          & 83.77 & 83.99 & 85.83 & 72.86 & 70.91 & 79.47 & 87.15          & 74.81          \\ \hline
T-patcher      & 1207 & 98.75 & 90.34          & 90.42 & 91.54 & 92.21 & 83.02 & 82.11 & 87.86 & \textbf{92.48} & \textbf{77.81} \\ \hline
MPN (only)     & 2519 & 99.20 & \textbf{94.33} & 94.97 & 95.34 & 95.34 & 86.10 & 82.85 & 92.92 & 92.26          & 77.06          \\ \hline
MPN (all) &
  4598 &
  \textbf{99.29} &
  93.05 &
  \textbf{98.39} &
  \textbf{98.45} &
  \textbf{98.58} &
  \textbf{95.53} &
  \textbf{95.30} &
  \textbf{96.94} &
  92.31 &
  77.04 \\ \hline
\end{tabular}
}
\label{Table1:XFEVER}
\end{table*}

\subsection{Overall Results}
\subsubsection{Results on XFEVER}
Table \ref{Table1:XFEVER} shows the experimental results on XFEVER dataset. Patch Num. denotes the number of added patches. Reliability denotes the accuracy on English editing set. MLG (monolingual generalization) denotes the generalization on English rephrase examples. CLG (cross-lingual generalization) denotes the generalization on cross-lingual parallel examples. Locality denotes the impact on unrelated English examples sampled from training set and test set. 

As shown in Table 1, the original model has a locality of 93.05\% on training set and 77.08\% on test set, with an accuracy of 88.07 on editing set. For other languages, the original model itself has cross-lingual generalization ability to some extent. This is primarily due to the shared representation space inherent in language models across different languages. When fine-tuning the mBERT model for specific downstream task using only English example, the resulting model retains ability to perform same task in other languages. This further validates the cross-lingual capabilities of language models. 

While for editing methods, we compared the following methods: 1) \textbf{Fine-tuning}, which edits the last FNN layer and uses cross-entropy loss as a constraint. It can be observed that simple fine-tuning cause suboptimal performance on various test metrics because of  forgetting problems. 2) \textbf{T-patcher}, which trains monolingual patch neurons with English examples and tests the generalization on other languages. The Reliability value of T-patcher is 98.75 and it can generalize to de, fr, and es, resulting in CLG values ranging from 0.90 to 0.92. This demonstrates that T-patcher also possesses some cross-lingual ability, but there is still a significant transfer gap. For languages from different families, such as ar and zh, their CLG values only increase from 0.8169/0.8273 to 0.8210/0.8302, indicating that generalization across different language families cannot be achieved. 3) \textbf{MPN}, which trains multilingual patch neuron using English example along with sampled example  from another language. When using the English example and the sampled example as conditions to judge whether the current input needs editing, i.e. MPN (only), the average CLG improvement is about 9\%. This suggests that the multilingual patch neuron trained in this manner has good cross-lingual generalization. When using examples from six language corresponding to the current input as conditions, i.e. MPN (all), the average CLG improvement increases to about 12\%. Furthermore, the MLG and Locality values of the MPN method are also satisfactory, indicating its effectiveness. 

\subsubsection{Results on XNLI}
For XNLI dataset, the experimental results are shown in Table \ref{Table2:XNLI}. It can be seen that the accuracy of the original model on editing set is 81.66, but its cross-lingual generalization effect is not ideal, with Thai (th) performing the worst due to the absence of th during mBERT's pre-training. The low-resource language sw also exhibits poor cross-lingual generalization which is only 49.82. For editing method, after editing with Fine-tuning, the cross-lingual generalization on editing set further decreases. T-patcher improves the cross-lingual generalization effect to some extent, but since its editing does not specifically optimize for cross-lingual characteristics, the improvement is limited, with only a 1\% increase in average CLG value compared to the original model. However, for MPN, there is an increase of 14.07\% on average CLG than that of the original model. This indicates that MPN can significantly improve the cross-lingual editing effect.

\begin{table*}
\caption{Results on XNLI dataset}
\resizebox{\linewidth}{!}{
\begin{tabular}{ccccccccccccccccccc}
\hline
\multirow{2}{*}{\begin{tabular}[c]{@{}c@{}}Models/\\ Methods\end{tabular}} &
  \multirow{2}{*}{Patch Num.} &
  \multirow{2}{*}{\begin{tabular}[c]{@{}c@{}}Reliability\\ (en)\end{tabular}} &
  \multicolumn{15}{c}{CLG} &
  \multirow{2}{*}{Locality} \\ \cline{4-18}
 &
   &
   &
  ar &
  bg &
  de &
  el &
  es &
  fr &
  hi &
  ru &
  sw &
  th &
  tr &
  ur &
  vi &
  zh &
  Avg. &
   \\ \hline
Original model &
  0 &
  81.66 &
  63.75 &
  69.90 &
  70.92 &
  66.95 &
  74.53 &
  72.61 &
  61.00 &
  69.38 &
  49.82 &
  34.95 &
  62.24 &
  58.50 &
  66.21 &
  66.11 &
  63.43 &
  81.81 \\ \hline
Fine-tuning &
  0 &
  75.79 &
  56.03 &
  61.04 &
  63.69 &
  58.88 &
  68.00 &
  65.37 &
  53.99 &
  60.68 &
  46.07 &
  35.35 &
  54.19 &
  52.14 &
  58.98 &
  55.99 &
  56.46 &
  76.51 \\ \hline
T-patcher &
  954 &
  \textbf{99.66} &
  64.33 &
  71.44 &
  73.71 &
  67.70 &
  79.44 &
  76.21 &
  61.32 &
  70.68 &
  50.08 &
  35.11 &
  62.59 &
  58.82 &
  66.91 &
  66.27 &
  64.62 &
  \textbf{81.65} \\ \hline
MPN &
  2254 &
  99.40 &
  \textbf{78.86} &
  \textbf{86.13} &
  \textbf{87.56} &
  \textbf{82.91} &
  \textbf{90.82} &
  \textbf{89.46} &
  \textbf{75.39} &
  \textbf{84.99} &
  \textbf{58.04} &
  \textbf{35.67} &
  \textbf{76.59} &
  \textbf{71.88} &
  \textbf{80.64} &
  \textbf{81.22} &
  \textbf{77.15} &
  81.16 \\ \hline
\end{tabular}
}
\label{Table2:XNLI}
\end{table*}

\subsection{Discussion}
To directly observe the generalization effect of editing methods on multiple languages, we collect 722 examples from XFEVER editing set across six languages where the original model made incorrect predictions. These examples formed the editing set. The results are presented in Table \ref{Table3:error_dataset}. It is evident that while Fine-tuning performs well on these examples, it significantly disrupts the locality of both the training and test sets. T-patcher exhibits improved locality but falls short in achieving cross-lingual generalization, particularly for zh and ar, where the edited knowledge does not generalize effectively. On the other hand, the proposed MPN demonstrates strong cross-lingual generalization as well as good locality.

\begin{table*}
\caption{Results on the set of examples where predictions in six languages are incorrect on XFEVER editing set.}
\resizebox{\linewidth}{!}{
\begin{tabular}{cccccccccccc}
\hline
\multirow{2}{*}{Models} &
  \multirow{2}{*}{\begin{tabular}[c]{@{}c@{}}Patch\\ Num.\end{tabular}} &
  \multirow{2}{*}{Reliability} &
  \multirow{2}{*}{MLG} &
  \multicolumn{6}{c}{CLG} &
  \multicolumn{2}{c}{Locality} \\ \cline{5-12} 
               &     &                &       & de    & fr    & es    & zh    & ar    & Avg.  & Train & Test           \\ \hline
Original model & 0   & 0              & 0     & 0     & 0     & 0     & 0     & 0     & 0     & 93.05 & 77.08          \\ \hline
Fine-tuning    & 45  & 99.45          & 96.72 & 96.95 & 94.74 & 95.43 & 98.61 & 96.12 & 96.37 & 9.85  & 26.26          \\ \hline
T-patcher      & 710 & \textbf{97.65} & 66.26 & 57.62 & 64.96 & 68.84 & 5.96  & 10.80 & 41.64 & 92.36 & \textbf{77.10} \\ \hline
MPN &
  711 &
  97.51 &
  \textbf{75.73} &
  \textbf{98.89} &
  \textbf{99.58} &
  \textbf{99.58} &
  \textbf{98.89} &
  \textbf{99.17} &
  \textbf{99.22} &
  \textbf{92.43} &
  77.03 \\ \hline
\end{tabular}
}
\label{Table3:error_dataset}
\end{table*}

\section{Conclusion}
Model editing is able to modify the knowledge stored in large models accurately without affecting other model behaviors. This paper explores the cross-lingual generalization of model editing and introduces a method for cross-lingual editing using multilingual patch neurons. This approach minimally alters the original editing method by incorporating cross-lingual sampling at the input to enhance its cross-lingual editing capability. Our findings suggest that the fine-tuning based editing methods, such as Fine-tuning and Transformer-patcher, demonstrate some degree of cross-lingual generalization. This capability mainly arises from the cross-lingual transferability of multilingual models rather than the editing methods. To this end, this study trains multilingual patch neurons through cross-lingual sampling at the input of editing methods, resulting in better cross-lingual editing result. 

\textbf{Limitations}: In this work, the cross-lingual editing method is only applied to classification tasks, specifically the XFEVER and XNLI datasets, without addressing more complex tasks such as question-answering on zsRE dataset due to challenges in obtaining parallel corpora. Future work aims to utilize the Bi-zsRE dataset by \cite{wang2023cross}  to assess the cross-lingual editing performance on question-answering tasks, which present greater difficulties. Additionally, the FEVER dataset contains a large amount of knowledge-intensive data, including named entities such as people, places, and movie names, but the XFEVER dataset created in this work is obtained using machine translation models without undergoing further manual verification due to resource constraints. For future research, larger models like GPT-4 may be considered to potentially improve translation results.

\bibliography{abs}
\bibliographystyle{acl_natbib}

\appendix

\section{Implement Details}
\label{sec:appendix1}

\subsection{XFEVER Dataset}
\label{sec:appendix2}
To construct a cross-lingual model editing dataset, we translate the editing set of FEVER dataset used in \cite{huang2023transformer,de2021editing} into five languages: de, fr, es, zh, and ar. We use the Helsinki-NLP machine translation models from Huggingface with the following models: opus-mt-en-de, opus-mt-en-fr, opus-mt-en-es, opus-mt-en-zh, and opus-mt-en-ar. The translated field is "input", and the resulting dataset is called XFEVER, which contains six languages: en, de, fr, es, zh, and ar, covering three different language families. The average sentence length in above six languages of XFEVER is 8.2, 7.8, 8.9, 8.8, 13.2, and 7.4. The distribution of sentence length is shown in the figure 3. The hyperparameters for training the initial model are set as shown in Table \ref{Table4:Datasets}. We select the model that performs best on the validation set, with an accuracy of 88.11\%.

\subsection{XNLI Dataset}
\label{sec:appendix3}
We fine-tune the mBERT model on MultiNLI training set (a total of 392,702 examples) and use the English version of XNLI development set (a total of 2490 examples) as the criterion for selecting the best model. Details of XNLI dataset and hyperparameters settings for training the original model is listed in Table \ref{Table4:Datasets}. We set the batch size to 64 and the epoch to 3. Finally, we select the best model with an accuracy of 0.750 on training set and 0.828 on development set. This accuracy is comparable to the official report of the mBERT model. 

\begin{figure*}
    \centering
    \includegraphics[width=0.3\linewidth]{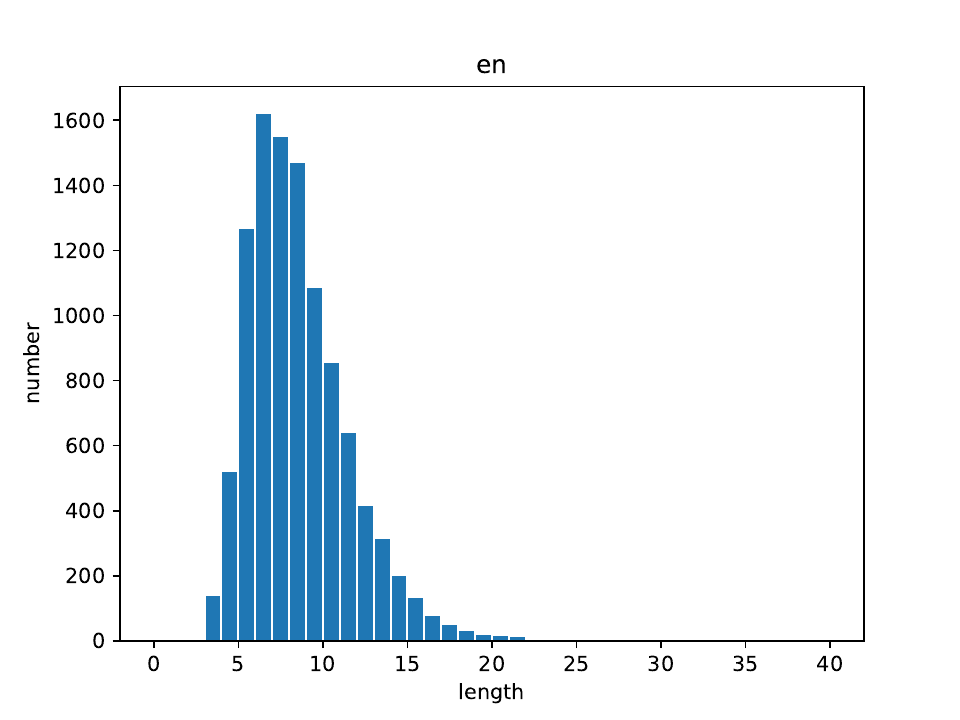}
    \includegraphics[width=0.3\linewidth]{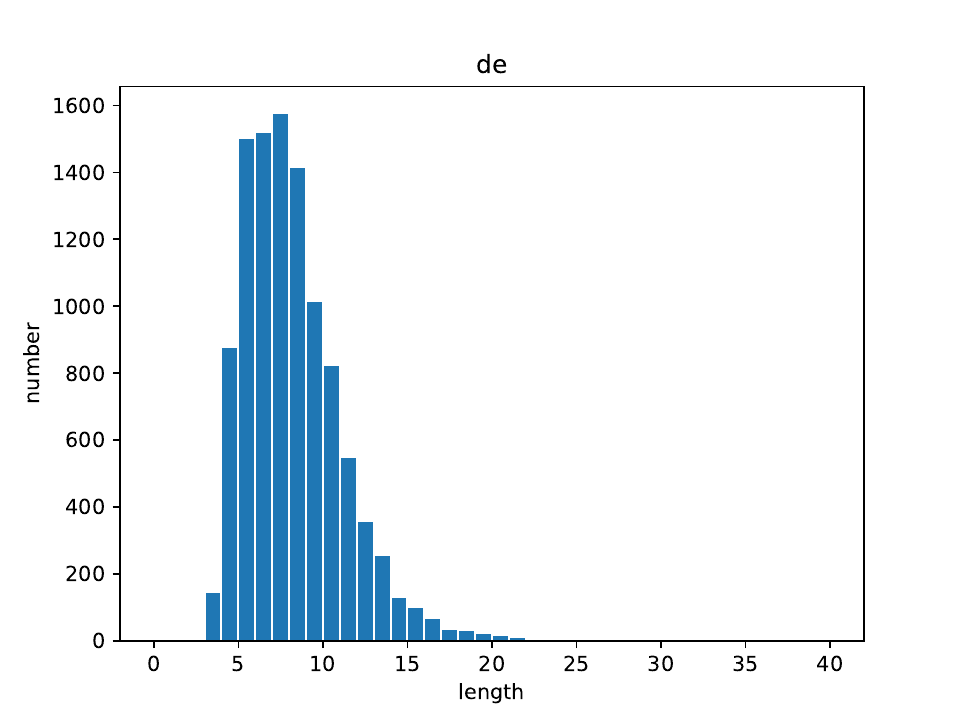}
    \includegraphics[width=0.3\linewidth]{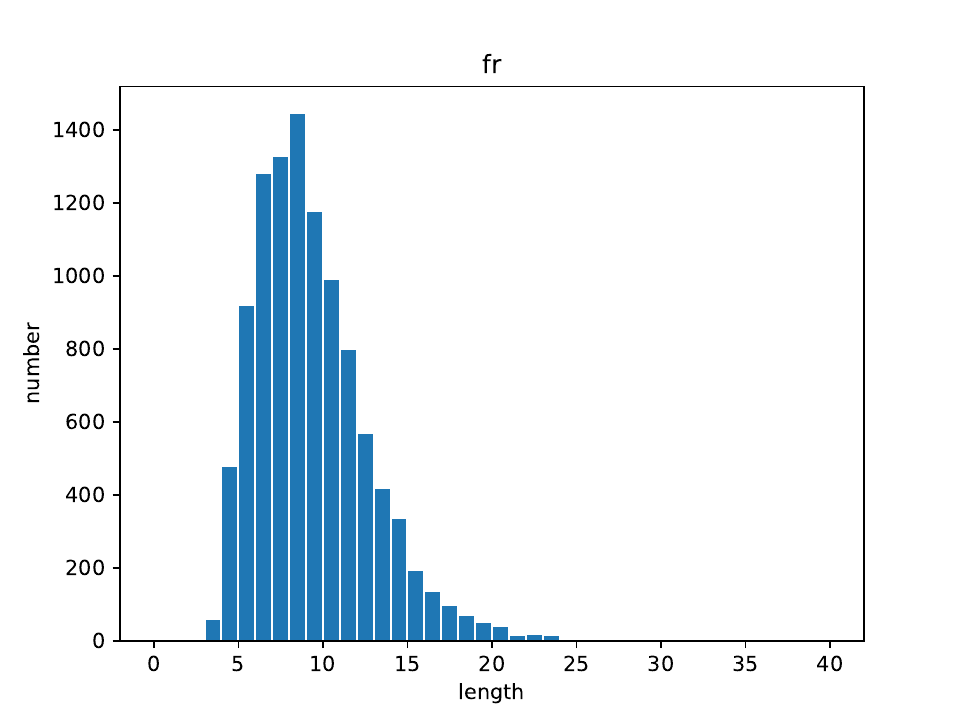}
    \includegraphics[width=0.3\linewidth]{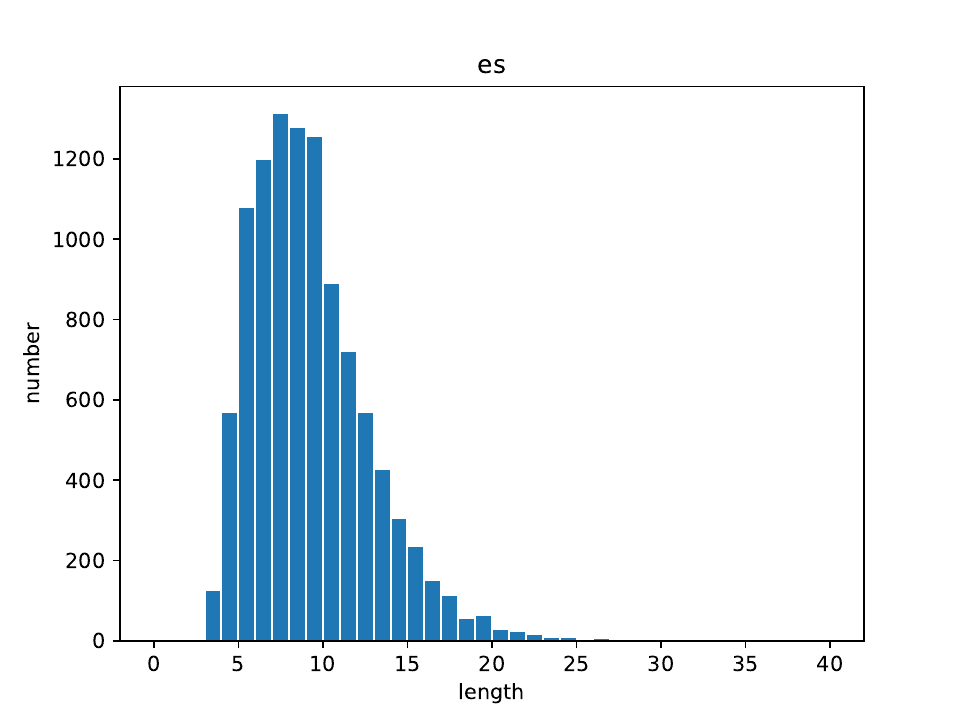}
    \includegraphics[width=0.3\linewidth]{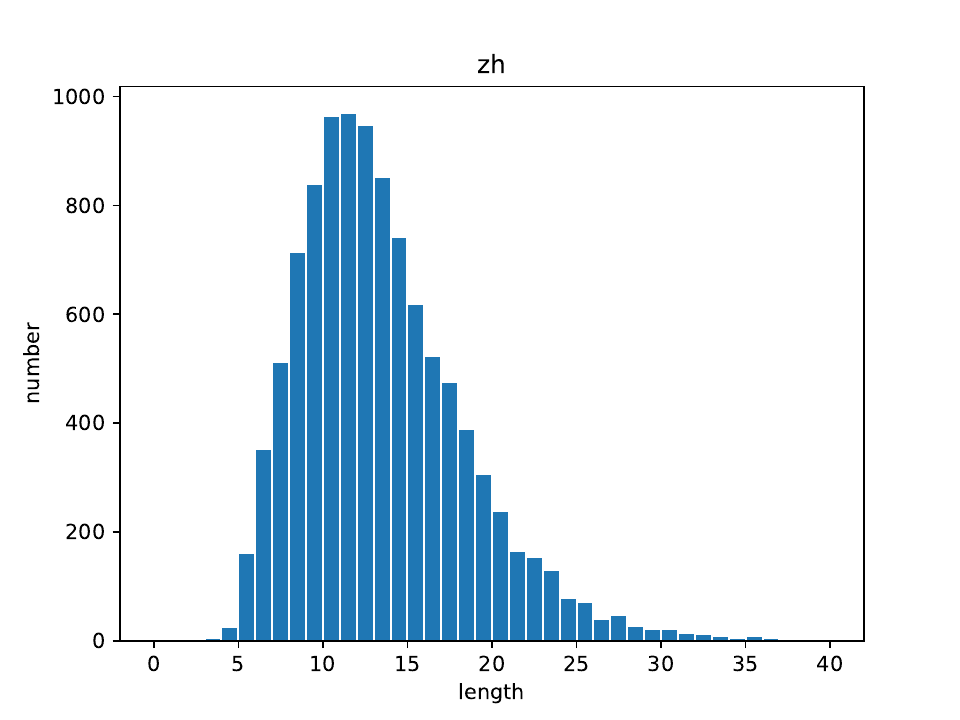}
    \includegraphics[width=0.3\linewidth]{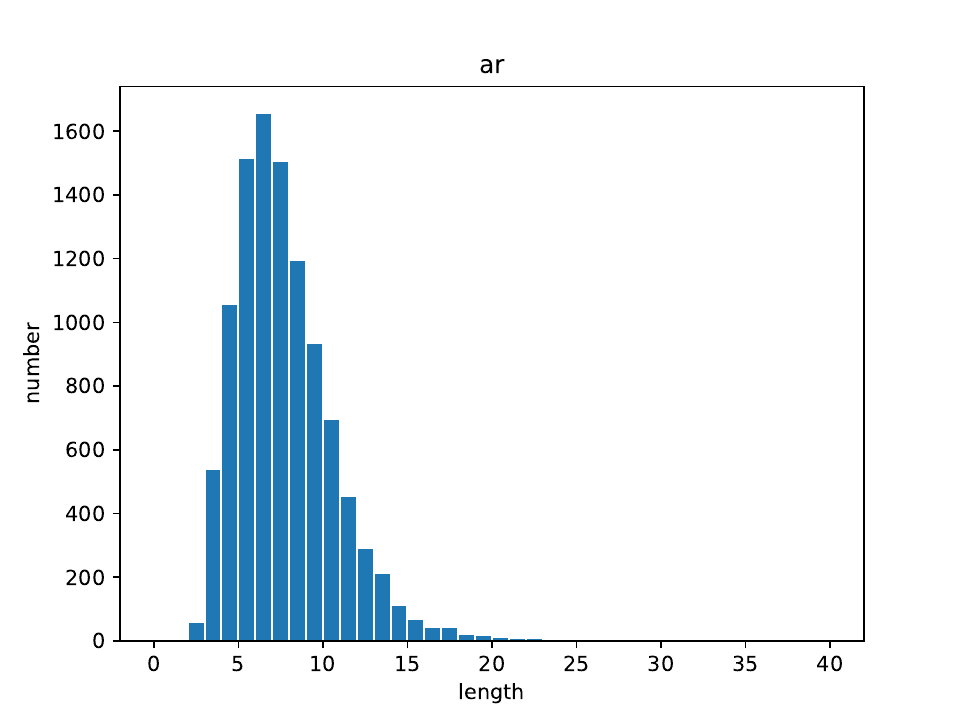}
    \caption{The distribution of sentence length for six languages on XFEVER dataset}
\end{figure*}

\begin{table*}
\caption{Datasets and hyperparameter settings for training the original model on XFEVER and XNLI}
\centering
\begin{tabular}{|c|c|c|}
\hline
Names           & XFEVER & XNLI    \\ \hline
train set       & 83,972 & 392,702 \\ \hline
validation set  & 10,496 & 2490    \\ \hline
editing set     & 10,496 & 5010    \\ \hline
test set        & 10,444 & --      \\ \hline
epoch           & 20     & 3       \\ \hline
batch\_size     & 128    & 64      \\ \hline
learning rate   & 3e-5   & 3e-5    \\ \hline
max\_length     & 32     & 128     \\ \hline
weight\_decay   & 0.01   & 0.01    \\ \hline
warmup\_updates & 500    & 500     \\ \hline
eps             & 0.1    & 0.1     \\ \hline
\end{tabular}
\label{Table4:Datasets}
\end{table*}

\end{document}